\documentclass{article}
\usepackage{spconf,amsmath,graphicx}
\usepackage{txfonts}
\usepackage{graphicx}
\usepackage{amsmath,amssymb}
\usepackage{url}
\usepackage[square, comma, sort&compress, numbers]{natbib}
\usepackage{subfigure}

\title{HIGH ORDER STRUCTURE DESCRIPTORS FOR SCENE IMAGES}
\name{Wenya Zhu$^1$, Xiankai Lu$^1$, Tao Xu$^2$, Ziyi Zhao$^1$ \thanks{Corresponding author: carrierlxk@gmail.com}}
\address{$^1$Department  of Automation, Shanghai Jiao Tong University, Shanghai, China\\
         $^{2}$School  of Aeronautics and Astronautics, Shanghai Jiao Tong University, Shanghai, China}
\begin{document}
\ninept
\bibliographystyle{unsrt}
\maketitle
\begin{abstract}
Structure information is ubiquitous in natural scene images and it plays an important role in scene representation. In this paper, third order structure statistics (TOSS) and fourth order structure statistics (FOSS) are exploited to encode higher order structure information. Afterwards,  based on the radial and normal slice of TOSS and FOSS, we propose the high order structure feature: third order structure feature (TOSF) and fourth order structure feature (FOSF). It is well known that scene images are well characterized by particular arrangements of their local structures, we divide the scene image into the non-overlapping sub-regions and compute the proposed higher order structural features among them. Then a scene classification is performed by using SVM classifier with these higher order structure features. The experimental results show that higher order structure statistics can deliver image structure information well and its spatial envelope has strong discriminative ability.
\end{abstract}
\section{INTRODUCTION}
\label{sec:intro}
Scene presentation is a basic issue towards scene categorization. Many features descriptors have been used for scene representation during last decade, including local descriptors, like SIFT~\cite{C26}, HOG~\cite{C27}, LBP~\cite{C25} $\emph{et.al.}$  and holistic descriptors, like Gist~\cite{C1,C28} $\emph{et.al}$.
Scene images have rich structure information which plays an important role in scene representation ~\cite{C1,C2,C20}. Several previous studies have concluded that the phase spectrum in the Fourier domain carries the majority of image structure information ~\cite{C3,C4}. However, it also suffers the widespread structural redundancy and phase sensitivity to variation ~\cite{C3,C5}. Therefore, it is necessary to find some appropriate ways to represent structural information. Higher order spectrum carries the Fourier phase spectrum, that is, structural information can be represented based on higher order statistics ~\cite{C6,C7}. In addition, existing psychophysical experiments have shown that discrimination performance is related to changes in higher-order image statistics ~\cite{C3,C6}.

As higher order statistics have the immunity to additive Gaussian noise and have translation invariance, several early studies used higher order cumulants and polyspectra to extract features ~\cite{C8,C9,C10,C13,C14,C15}. One way is the matched filtering in the cumulant domain ~\cite{C13,C14}. Tsatsanis and Giannakis  ~\cite{C14} classified object and texture by exploiting the minimum-distance object classifier based on matching filtering in the third order cumulant domain. However, this method needs a large amount of computation. In ~\cite{C8,C9,C10}, integrated bispectral-based feature extraction methods are proposed to solve this problem. The vital information for image identification may be lost since only some slices are utilized. Furthermore, it is not certain whether the chosen slices as feature vectors have the discriminant power. In order to avoid blind slices, the bi-spectra set with most discriminant power as feature vectors at individual frequency points are selected ~\cite{C15}.

In this paper, we propose higher order structure statistics as a novel structure information descriptor of scene images and  we use the proposed descriptor to extract scene features. To the best of our knowledge, higher order statistics are seldom applied to scene structure information representation. 

Motivated by the relationships between phase spectrum, bi-spectrum and tri-spectrum, we derive the third order structure statistics (TOSS) and fourth order structure statistics (FOSS). And inspired by the spatial envelope of the scene representation ~\cite{C1}, we regard a scene as an individual object with unitary surface pictures of TOSS and FOSS. Third order structure feature (TOSF) and fourth order structure feature (FOSF) are then calculated based on the radical and normal slice of TOSS  and the normal slice of FOSS, respectively. Considering scene images are well characterized by particular arrangements of their local structures ~\cite{C17}, we divide image into the non-overlapping sub-regions. Then proposed higher order structural features among all sub-regions are computed. Finally, the mean, variation and energy are further computed from these higher order structure features of all sub-regions to form the feature vectors. Subsequently, scene categorization is adopted to validate the effectiveness of high order structure features.

\section{THIRD ORDER STRUCTURE FEATURE (TOSF) AND FOURTH ORDER STRUCTURE (FOSF)}
\label{sec:format}
In this section, we briefly introduce the definition of bi-spectrum and tri-spectrum. Then, third order structure statistics (TOSS) and fourth order structure statistics(FOSS) are developed as two novel structure information descriptors.
\subsection{The definition of the higher order spectrum}
Higher order spectrum (also known as poly-spectra) is defined as the discrete-time Fourier transform of the corresponding higher order cumulant.
The bi-spectrum and tri-spectrum are special cases of the $\emph{n}$-th order poly-spectrum ~\cite{C19}. The bi-spectrum of one image is:
\begin{equation}
\begin{split}
{B_f}({\omega _{x1}},{\omega _{y1}},{\omega _{x2}},{\omega _{y2}}) = &F({\omega _{x1}},{\omega _{y1}})F({\omega _{x2}},{\omega _{y2}})\\
&{F^ * }({\omega _{x1}} + {\omega _{x2}},{\omega _{y1}} + {\omega _{y2}})\
\end{split}
\end{equation}
where $F({\omega _x},{\omega _y})$ is the Fourier transform of the image $f(m,n)$.  Eq. 1 is a tri-nary product of Fourier coefficients and a complex-valued function with two frequency variables. Like the bi-spectrum, the tri-spectrum of the image
$f(m,n)$ is defined as the following quadruple product:
\begin{equation}
\begin{aligned}
T_f({\omega_{x1}},{\omega_{y1}},{\omega_{x2}},{\omega_{y2}},{\omega_{x3}},{\omega _{y3}}) = F({\omega_{x1}},{\omega_{y1}})F({\omega_{x2}},{\omega_{y2}})&\\
F({\omega_{x3}},{\omega_{y3}})F^ *({\omega _{x1}} + {\omega _{x2}} + {\omega _{x3}},{\omega _{y1}} + {\omega _{y2}} + {\omega _{y3}})&
\end{aligned}
\end{equation}
with six frequency variables. Clearly, the poly-spectra with the higher order than fourth have more frequency variables, which means that there is a high-computational complexity. In this paper, we focus on three order and four order spectrum.
\subsection{Higher order structure statistics}
\subsubsection{Third order structure statistics(TOSS)}
The higher-order statistics have been widely applied in many diverse fields ~\cite{C7}. However, the high dimension of poly-spectra impedes investigating higher-order statistics of scene images. In addition, there are widespread structural redundancy and phase sensitivity to variation ~\cite{C5,C3,C5}. It implies that the poly-spectra cannot be utilized directly and effectively. Since higher order spectrum carries the Fourier phase spectrum, the relationship between poly-spectra and phase spectra is very important to establish higher order structure statistics and represent structure information appropriately.

The Brillinger algorithm is a recursive approach to reconstruct phase spectra from the bi-spectrum [19]. To obtain the relationship between poly-spectra and phase spectra, we extend the Brillinger algorithm from 1D  signal to 2D image.
We have seen that the bi-spectrum of the image is given by Eq. 2. The Fourier transform and bi-spectrum in plural form can be rewritten as:
\begin{equation*}
\begin{aligned}
&F({\omega _x},{\omega _y}) = \left| {F({\omega _x},{\omega _y})} \right|\exp [ - j\varphi ({\omega _x},{\omega _y})]\\
&{B_f}({\omega _{x1}},{\omega _{y1}},{\omega _{x2}},{\omega _{y2}})  = \left| {{B_f}({\omega _{x1}},{\omega _{y1}},{\omega _{x2}},{\omega _{y2}})} \right|\\
&\exp [ - j\psi ({\omega _{x1}},{\omega _{y1}},{\omega _{x2}},{\omega _{y2}})]\\
\end{aligned}
\end{equation*}
Combining Eq. 1 with the equations above, we obtain:
\begin{equation}
\begin{aligned}
\psi ({\omega _{x1}},{\omega _{y1}},{\omega _{x2}},{\omega _{y2}}) = &\varphi ({\omega _{x1}},{\omega _{y1}}) + \varphi ({\omega _{x2}},{\omega _{y2}})\\
&-\varphi ({\omega _{x1}} + {\omega _{x2}},{\omega _{y1}} + {\omega _{y2}})
\end{aligned}
\end{equation}
where $\psi ({\omega _{1x}},{\omega _{1y}},{\omega _{2x}},{\omega _{2y}})$ is recognized as the phase of the bi-spectrum and $\varphi ({\omega _x},{\omega _y})$ is the image phase spectra. Summation of Eq. 3 with the constrains that
\begin{equation}
\begin{aligned}
&{\omega _{x1}} + {\omega _{x2}} = {\omega _x}
{}&{(0 \le {\omega _{x1}} \le {\omega _x};0 \le {\omega _{x2}} \le {\omega _x})}
\\
&{\omega _{y1}} + {\omega _{y2}} = {\omega _y}
{}&{(0 \le {\omega _{y1}} \le {\omega _y};0 \le {\omega _{y2}} \le {\omega _y})}
\end{aligned}
\end{equation}
yields,
\begin{equation}
\begin{aligned}
\sum\limits_{\omega _x} \sum\limits_{\omega _y} \psi ({\omega _{x1}},{\omega _{y1}},{\omega _{x2}},{\omega _{y2}})    = &\sum\limits_{\omega _x} \sum\limits_{\omega _y} [\varphi ({\omega _{x1}},{\omega _{y1}}) + \varphi ({\omega _{x2}},{\omega _{y2}}) \\&-\varphi({\omega _{x1}} + {\omega _{x2}},{\omega _{y1}} + {\omega _{y2}})]
\end{aligned}
\end{equation}
Further, summing over ${\omega _{x1}} = [0,{\omega _x}]$, ${\omega _{y1}} = [0,{\omega _y}]$ and letting ${\omega _{x2}} = {\omega _x} - {\omega _{x1}}$, ${\omega _{y2}} = {\omega _y} - {\omega _{y1}}$, gives
\begin{equation}
\begin{aligned}
\sum\limits_{\omega _{x1} = 0}^{\omega_x} \sum\limits_{\omega _{y1} = 0}^{\omega _y} \psi ({\omega _{x1}},{\omega _{y1}},{\omega _x} - {\omega _{x1}},{\omega_y} - {\omega_{y1}})  = \sum\limits_{\omega _{x1} = 0}^{{\omega _x}} \sum\limits_{\omega _{y1} = 0}^{\omega _y} &\\
[\phi (\omega _{x1},\omega _{y1}) + \phi ({\omega _x} - {\omega _{x1}},{\omega _y} - {\omega _{y1}}) - \phi ({\omega _x},{\omega _y})]&
\end{aligned}
\end{equation}
For convenience, let $\Delta {\omega _x} = 1$, $\Delta {\omega _y} = 1$, $\omega _{x1} = {i_x}$, $\omega _{y1} = {i_y}$, ${\omega _x} = m$ and ${\omega _y} = n$. Eq. 6 can be reformulated as:
\begin{equation}
\begin{aligned}
&\sum\limits_{{i_x} = 0}^m {\sum\limits_{{i_y} = 0}^n {\psi ({i_x},{i_y},m - {i_x},n - {i_y})} }  = \\
&\sum\limits_{{i_x} = 0}^m {\sum\limits_{{i_y} = 0}^n {[\varphi ({i_x},{i_y}) + \varphi (m - {i_x},n - {i_y})] - (m + n + 2)\varphi (m,n)} }
\end{aligned}
\end{equation}
We notice that:
\begin{equation}
\sum\limits_{{i_x} = 0}^m {\sum\limits_{{i_y} = 0}^n {[\varphi ({i_x},{i_y}) + \varphi (m - {i_x},n - {i_y})]}  = 2} \sum\limits_{{i_x} = 0}^m {\sum\limits_{{i_y} = 0}^n {\varphi ({i_x},{i_y})} }
\end{equation}
Substituting Eq. 8 to Eq. 7  allows us rewrite Eq. 7 as
\begin{equation}
\begin{aligned}
\sum\limits_{{i_x} = 0}^m \sum\limits_{{i_y} = 0}^n \psi ({i_x},{i_y},m - {i_x},n - {i_y})  = &2\sum\limits_{{i_x} = 0}^{m - 1} \sum\limits_{{i_y} = 0}^{n - 1} \phi ({i_x},{i_y}) \\  & - (m + n) \phi (m,n)
\end{aligned}
\end{equation}
Therefore, Eq. 9 reflects the relationship between bi-spectra and phase spectra. To establish higher order structure statistics, we define
\begin{equation}
S(m,n) = \sum\limits_{{i_x} = 0}^m {\sum\limits_{{i_y} = 0}^n {\psi ({i_x},{i_y},m - {i_x},n - {i_y})}}
\end{equation}
and Eq. 9 can be written as:
\begin{equation}
\begin{aligned}
S(m,n) = & 2\sum\limits_{{i_x}= 0}^{m - 1} {\sum\limits_{{i_y} = 0}^{n - 1} {\phi ({i_x},{i_y})} -(m + n - 1)} \phi (m,n),\\
&{m = 1, \cdots ,M;n = 1, \cdots ,N}
\end{aligned}
\end{equation}
where $m = M,n = N$ corresponds to ${\omega _x} = \pi ,{\omega _y} = \pi $ respectively. It is obvious that $S(m,n)$ is the projection of $\psi ({\omega _{1x}},{\omega _{1y}},{\omega _{2x}},{\omega _{2y}})$ to the summation area. The projection achieves the dimensionality reduction of bi-spectrum. $S(m,n)$ is found to be only dependent on the phase spectra of the natural image. This suggests that $S(m,n)$ is capable of modelling the image structure information. Therefore, $S(m,n)$ can be defined as a novel structure descriptor, and called as third order structure statistics (TOSS). With the similar derivation process, we obtain fourth order structure statistics (FOSS) defined as follows:
\begin{equation}
\begin{aligned}
Q(m,n) =  3\sum\limits_{{k_x} = 0}^{m - 1} {\sum\limits_{{k_y} = 0}^{n - 1} {(m + 1 - {k_x})(n + 1 - {k_y})\phi ({k_x},{k_y})} } \\
 - \frac{{(m + 1)(m + 2)(n + 1)(n + 2)}}{4}\phi(m,n)
\end{aligned}
\end{equation}
Eq. 11 and Eq. 12 also provide a fast algorithm to directly compute TOSS and FOSS by linear combination of image phase spectrum, without first computing poly-spectra. It is found that the phase weighting coefficients in the Eq. 12 are determined by the location of the FOSS in the frequency domain. The variable weighting coefficients lead to encode the spatial distribution of the phase spectrum.

To verify effectiveness of higher order structure statistics which reflect image structure information, image reconstructions based on TOSS and FOSS are implemented respectively. Reconstructed image is obtained by inverse Fourier transform of TOSS and FOSS of original image with unity magnitude respectively~\cite{C10}.  It reveals that TOSS and FOSS indeed contain structure information and obtain better reconstruct performance than phase spectrum.\\

\begin{figure}[htbp]
\begin{flushleft}
\begin{minipage}[t]{0.48\linewidth}
  \centerline{\includegraphics[width=0.8\textwidth]{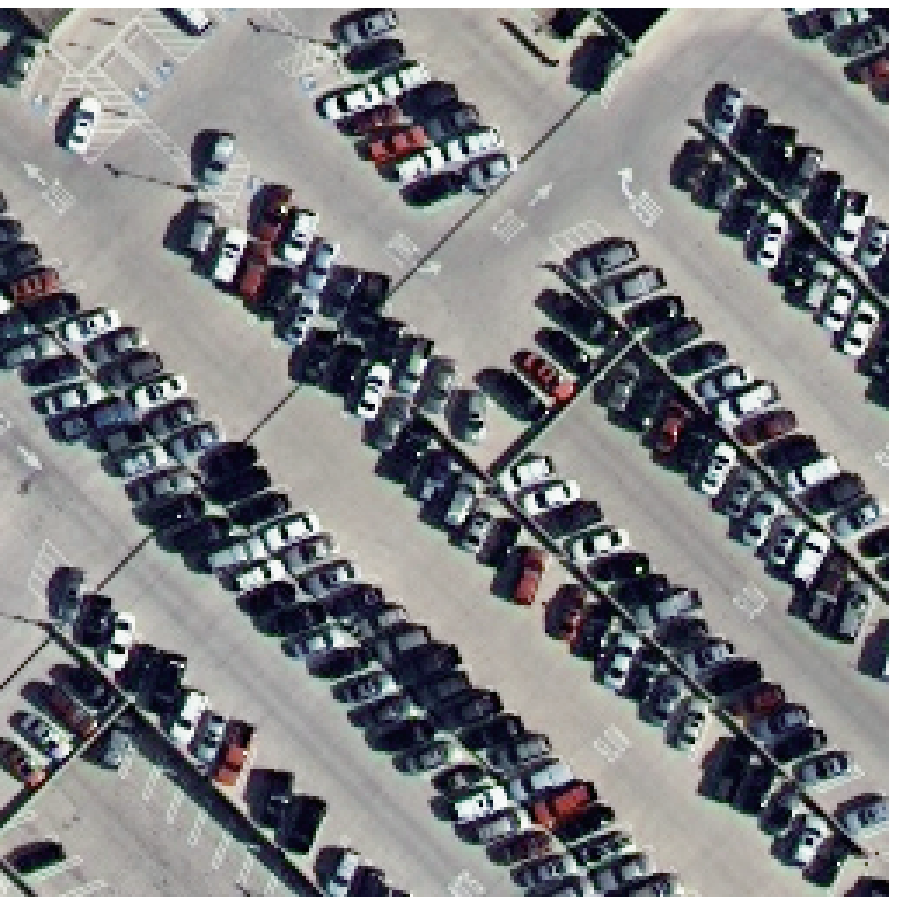}}
  \centerline{(a)}
\end{minipage}
\hfill
\begin{minipage}[t]{.48\linewidth}
  \centerline{\includegraphics[width=0.8\textwidth]{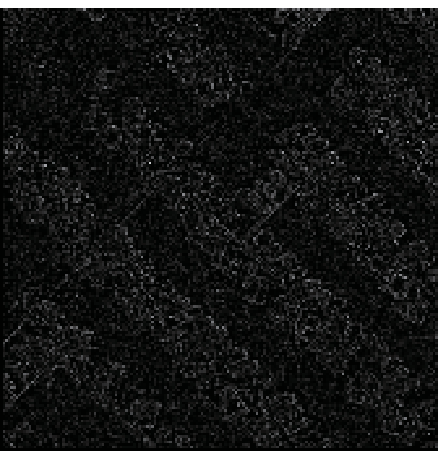}}
  \centerline{(b)}
\end{minipage}
\end{flushleft}
\begin{flushleft}
\begin{minipage}[t]{0.48\linewidth}
  \centerline{\includegraphics[width=0.8\textwidth]{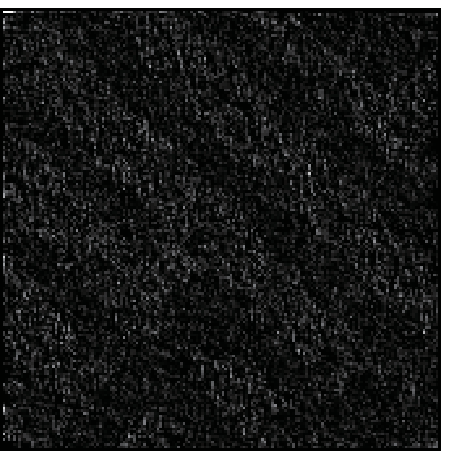}}
  \centerline{(c)}
\end{minipage}
\hfill
\begin{minipage}[t]{0.48\linewidth}
  \centerline{\includegraphics[width=0.8\textwidth]{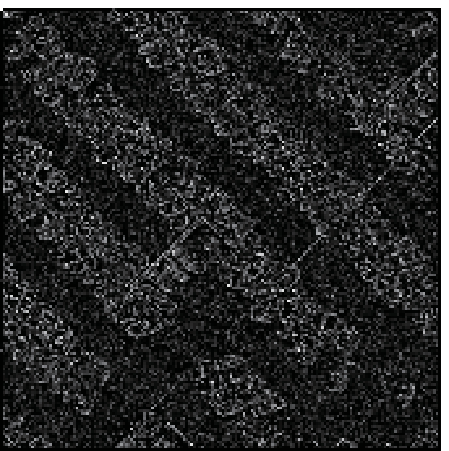}}
  \centerline{(d)}
\end{minipage}
\end{flushleft}
\vspace{-2em}
\caption{Image reconstruction based on phase, TOSS and FOSS. (a) Original image. (b) Image reconstructions based on phase spectrum. (c) Image reconstructions based on TOSS. (d) Image reconstructions based on FOSS.}
\vspace{-2em}
\label{fig:res}
\end{figure}
\subsection{TOSS and FOSS in different scene categories}
Also, we compute the TOSS and FOSS of different scene categories, like beach and agriculture (Fig.2(a)). Firstly, a log-polar transformation is first applied to two images to eliminate the influence of the rotation and scale. Then the mean normalization processing is used for eliminating differences in illumination. Finally, TOSS and FOSS of each scene category are calculated by averaging TOSS and FOSS over 100 transformed exemplars for each class, as shown in Fig.2. The surface representations of TOSS and FOSS show that different scene categories have different spatial envelopes. Therefore, these surface representations may be conceptualized in a unitary.
\begin{figure}[htb]
\begin{flushleft}
  \centerline{\includegraphics[width=0.18\textwidth]{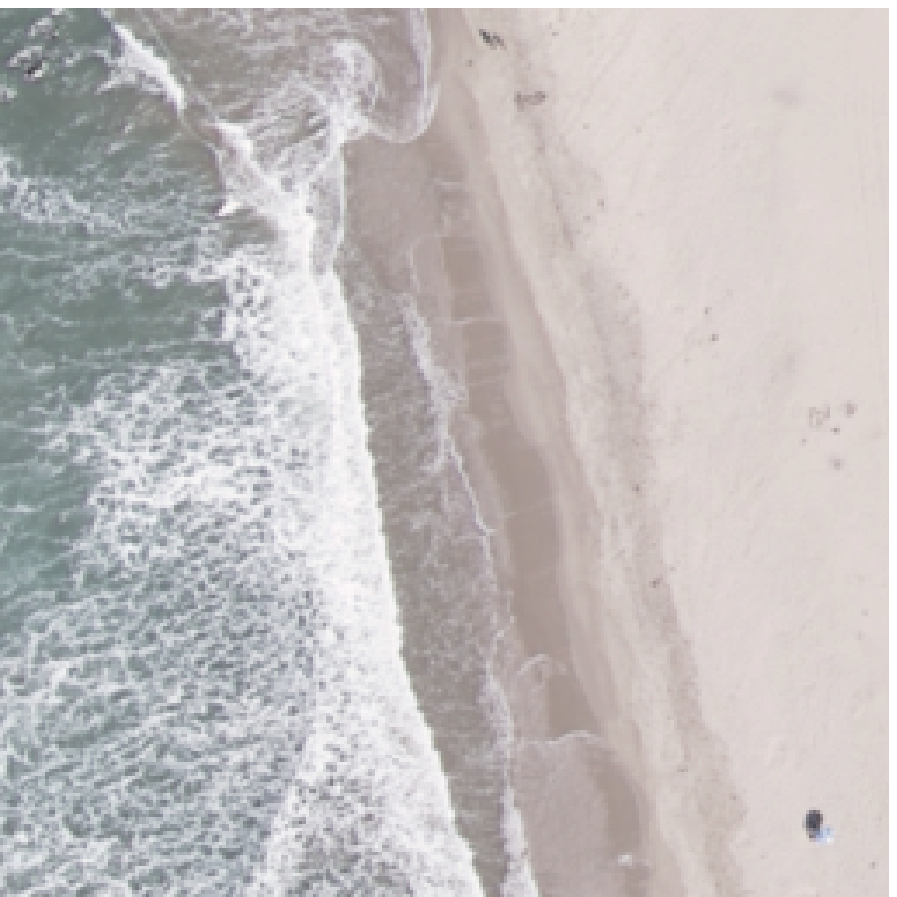}
  \includegraphics[width=0.18\textwidth]{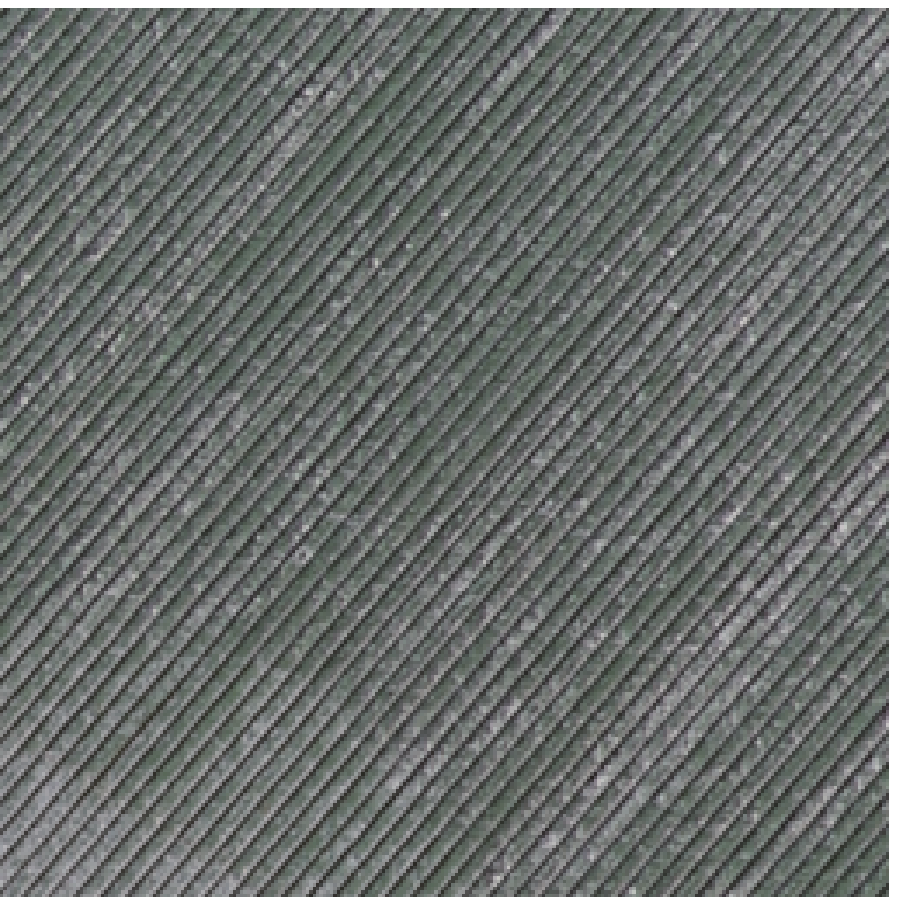}}
  \centerline{(a)}
\end{flushleft}
\vspace{-2em}
\begin{flushleft}
  \centerline{\includegraphics[width=0.2\textwidth]{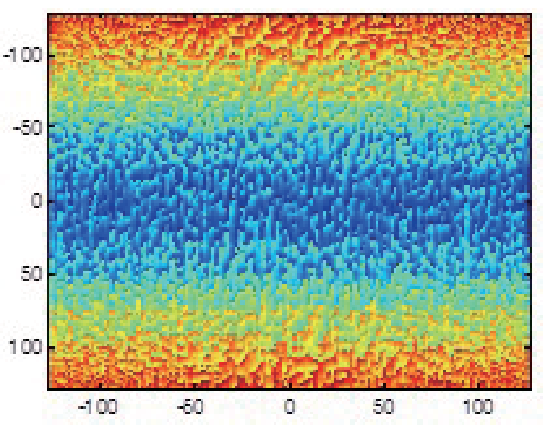}
  \includegraphics[width=0.21\textwidth]{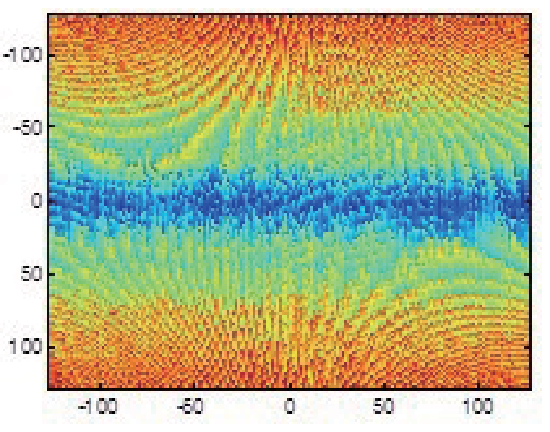}}
  \centerline{(b)}
\end{flushleft}
\vspace{-2em}
\begin{flushleft}
  \centerline{\includegraphics[width=0.2\textwidth]{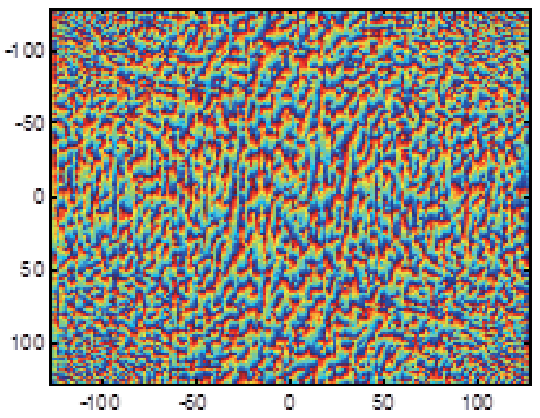}
  \includegraphics[width=0.2\textwidth]{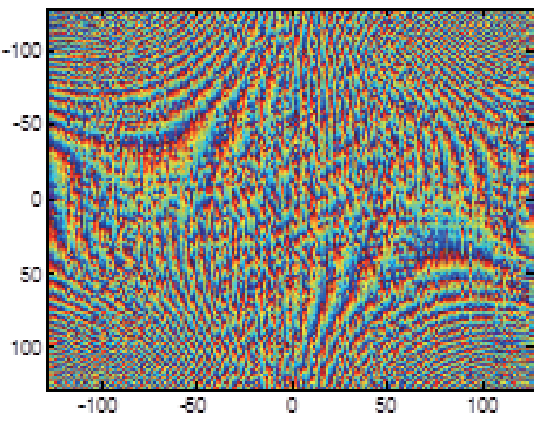}}
  \centerline{(c)}
\end{flushleft}
\vspace{-3em}
\caption{(a) Examples of scenes from beaches and agricultures (b) TOSS of beaches and agricultures (c) FOSS of beaches and agriculture.  For (b) and (c), the frequency  is located at center of each image.}
\vspace{-2em}
\label{fig:res}
\end{figure}

\section{SCENE CLASSIFICATION BASED ON HIGHER ORDER STRUCTURE STATISTICS}
\label{sec:pagestyle}
\subsection{Higher order structure feature}
As above discussed, the surface images of TOSS and FOSS may be conceptualized in a unitary form. Thus, higher order structure statistics can be suitable for extracting higher order structure features.
Fig. 2 has shown that the higher order statistics have categorization discrimination. TOSS and FOSS can be seen as the holistic features just like Gist~\cite{C1}. We extract the radial and normal slice of TOSS and FOSS and reduce them from two-dimension to one-dimension, as shown in Fig. 3. The radial and normal slices are obtained by averaging across horizontal and vertical orientations.
\begin{figure}[htb]
\begin{minipage}[b]{.48\linewidth}
  \centering
  \centerline{\includegraphics[width=4.0cm]{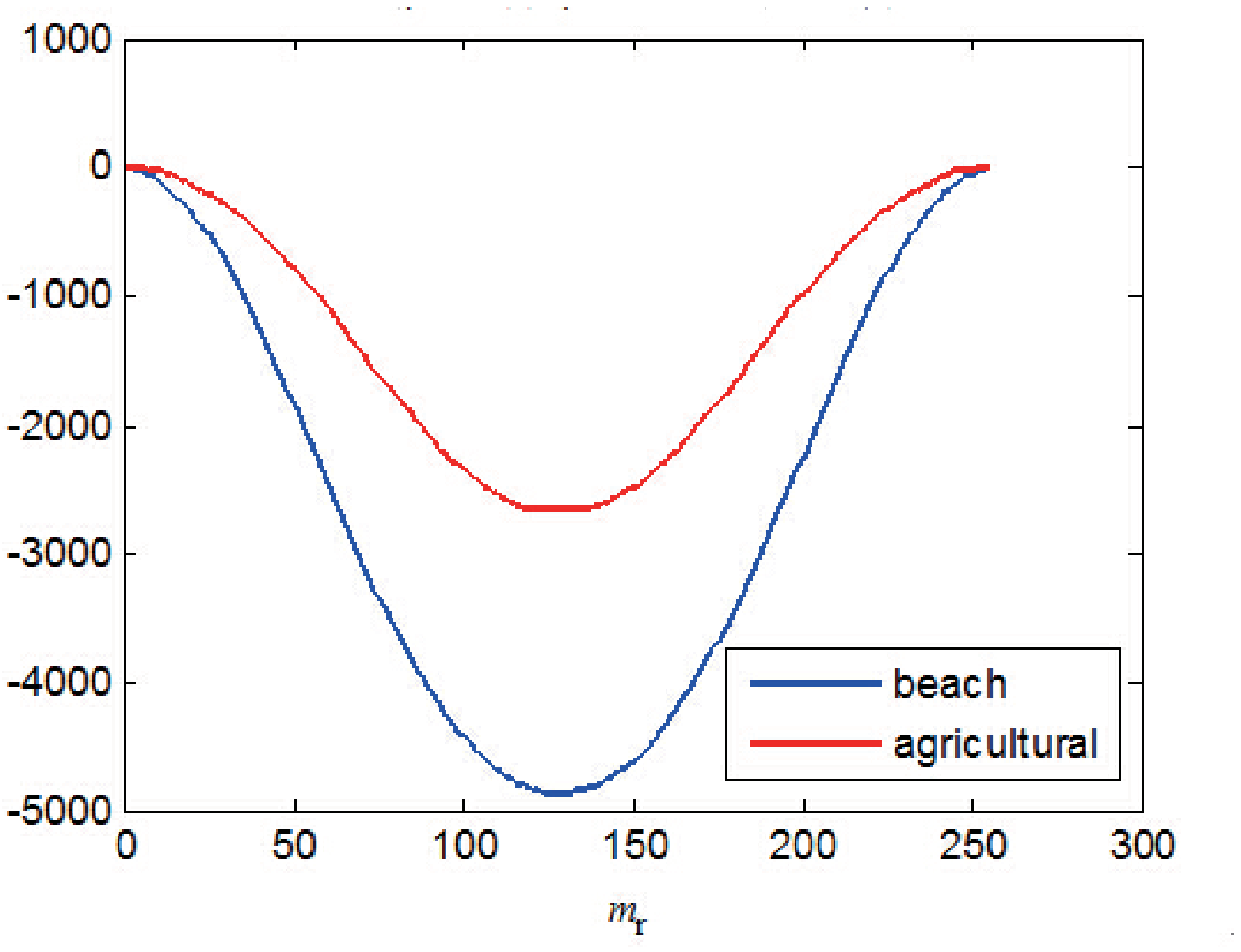}}
  \centerline{(a)}\medskip
\end{minipage}
\hfill
\begin{minipage}[b]{0.48\linewidth}
  \centering
  \centerline{\includegraphics[width=4.0cm]{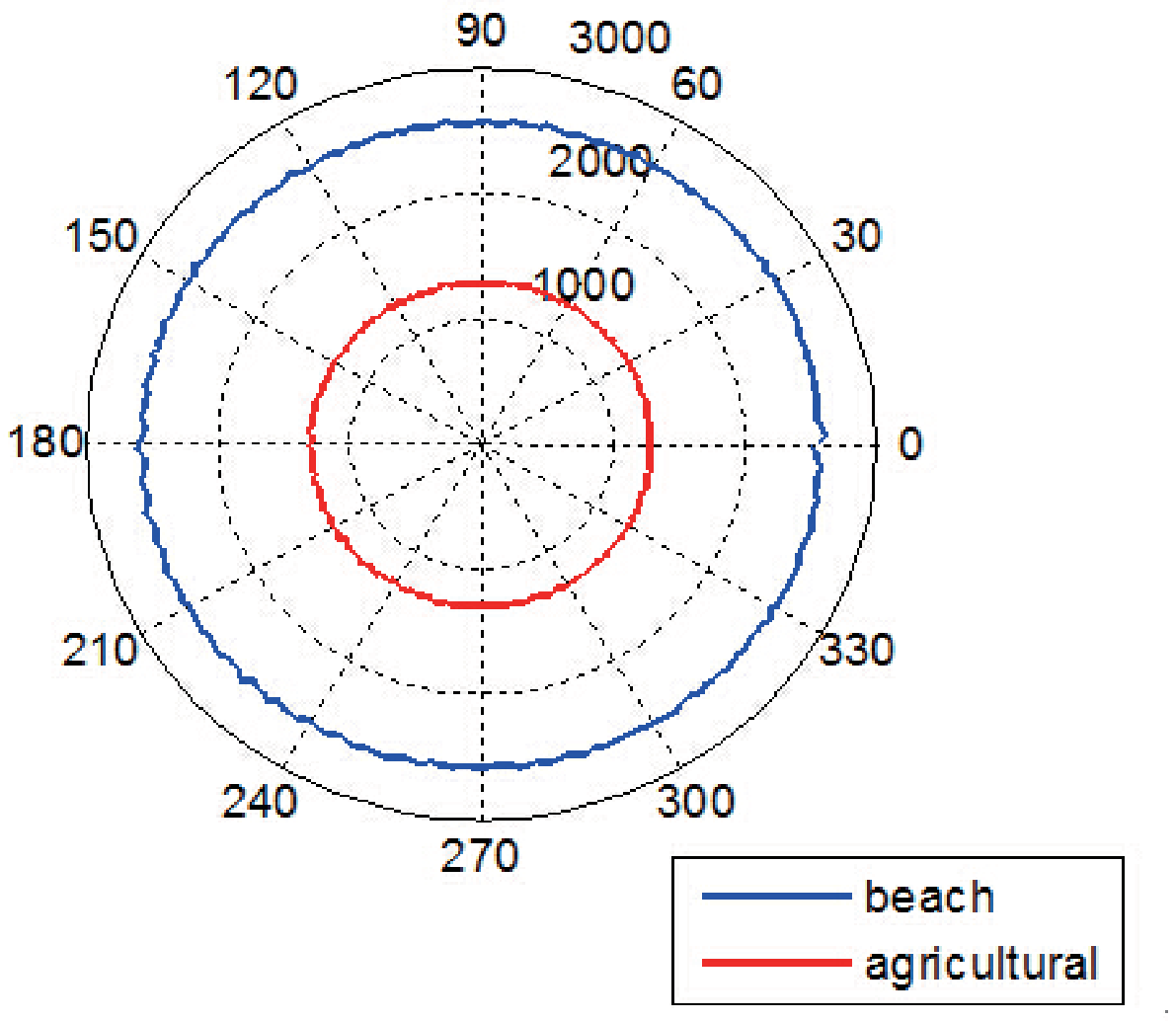}}
  \centerline{(b)}\medskip
\end{minipage}
\begin{minipage}[b]{1.0\linewidth}
  \centering
  \centerline{\includegraphics[width=6.0cm]{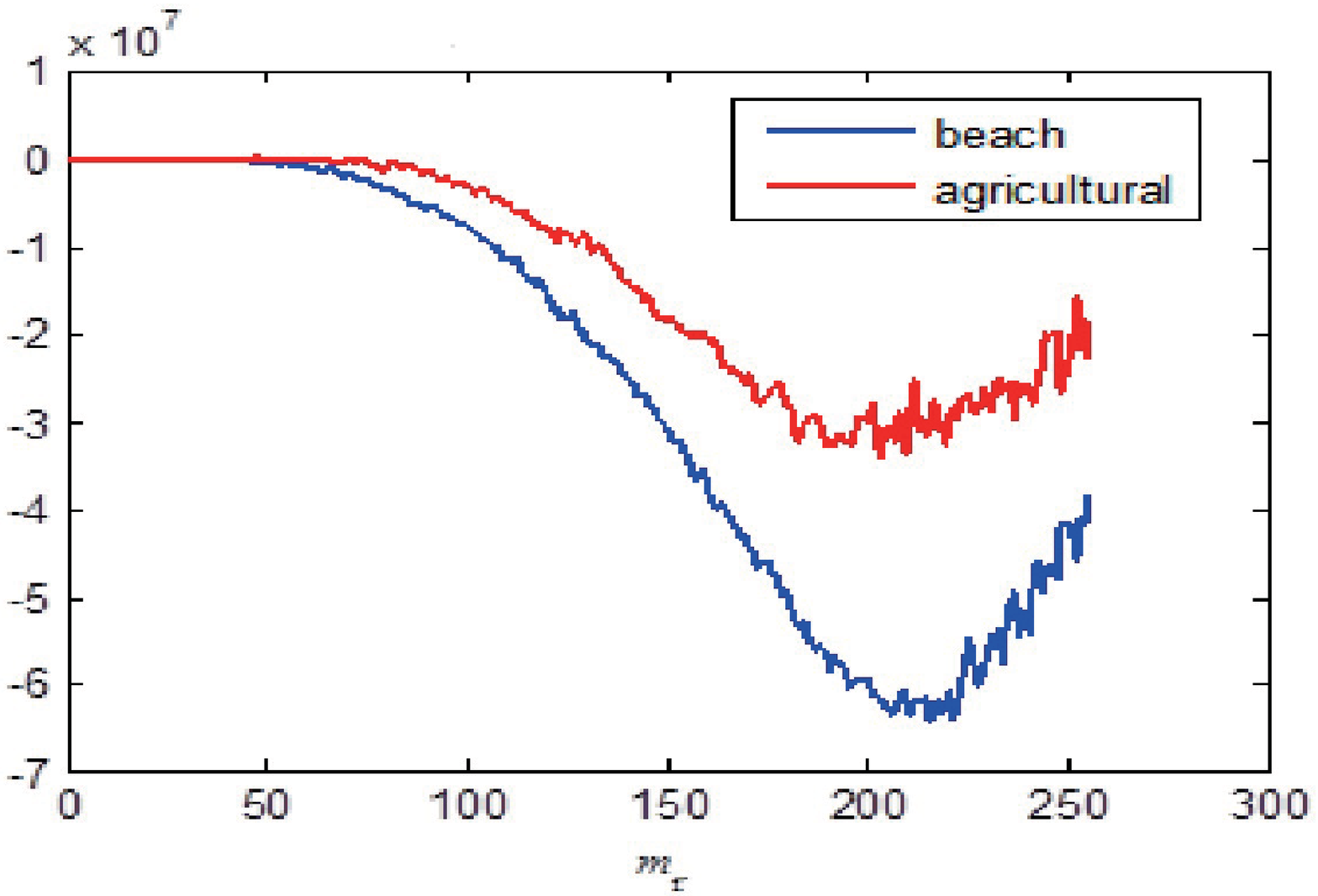}}
  \centerline{(c)}\medskip
\end{minipage}
\vspace{-2em}
\caption{The comparison of the TOSS radial slices. (a) the TOSS normal slices (b) the FOSS radial slices  (c) the FOSS normal slices. The normal slices is plotted in polar coordinate.}
\label{fig:res}
\end{figure}

TOSS radial slice is symmetric about origin point, but its minimum is different with various scene categories. Though TOSS normal slice mainly presents uniform distribution, TOSS normal slices of different scenes have different means. Therefore, the minimum of TOSS radial slice and the mean of TOSS normal slice can be used to characterize the spatial envelope attribute of TOSS. They are called third order structure feature (TOSF) in this paper. We notice that the TOSS values nearby zero frequencies have large fluctuations. To model TOSF more robustly, the mean of TOSS normal slice is replaced by the median. Formally, the TOSF of the image  $f(m,n)$ is formulated as:
\begin{equation}
TOSF(f(m,n)) = ({T_1},{T_2})
\end{equation}
where ${T_1}$ is the minimum of TOSS radial slice, and ${T_2}$  is the median of TOSS normal slice.
Similarly, the fourth order structure feature (FOSF) is defined as:
\begin{equation}
FOSF(f(m,n)) = ({F_1},{F_2})
\end{equation}
where ${F_1}$ is the minimum of FOSS radial slice, and ${F_2}$ is the value of FOSS normal slice at the maximum frequency. Since FOSS normal slice have no obviously consistent transformation regularity, this part of information is discarded.\\

\subsection{Experiments on higher order structure feature}
In this section, scene images classification is applied to validate the proposed higher order structure features. For scene images, it is well known that they are well characterized by particular arrangements of their local structures ~\cite{C2}. In Fig. 4, both forest and golf course have trees and grassland. The way to distinguish them is that the golf course has large continuous grassland, whereas the forest is dotted with grass. So the trees and grassland can be seen as the basic structure elements for forest and golf course scene and TOSF and FOSF can be used to represent these basic structure elements.
We use TOSF and FOSF in scene classification to verify their effectiveness and all images come from dataset ~\cite{C23}. 
\begin{figure}[htbp]
\begin{flushleft}
\begin{minipage}[b]{.48\linewidth}
  \centering
  \centerline{\includegraphics[width=3.8cm]{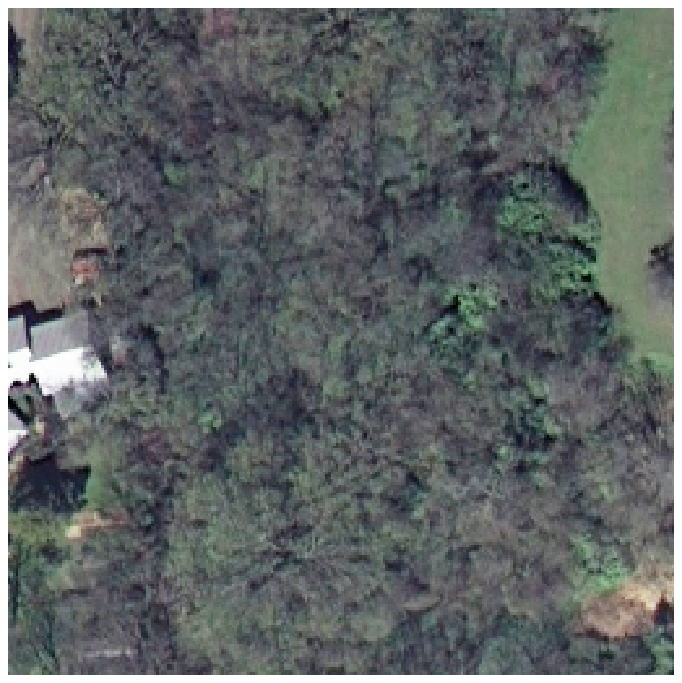}}
  \centerline{(a)}\medskip
\end{minipage}
\hfill
\begin{minipage}[b]{.48\linewidth}
  \centering
  \centerline{\includegraphics[width=3.8cm]{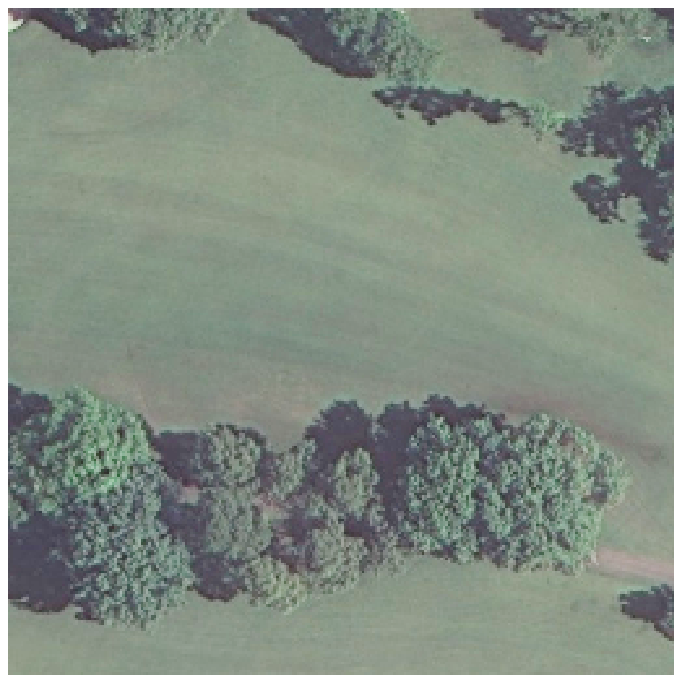}}
  \centerline{(b)}\medskip
\end{minipage}
\end{flushleft}
\vspace{-3em}
\caption{The image of the (a) forest and (b) golf course.}
\end{figure}

In order to robustly represent scene images and reflect the basic structure elements arrangement, the image is divided into the non-overlapping sub-regions. The advantage of this operation lies in the preservation of the arrangement of basic structure elements. Then the proposed higher order structural features among all sub-regions are computed. Finally, the mean, variation and energy are further calculated from these higher order structure features among all sub-regions as feature vectors. Then these feature vectors are fed into SVM classifier. For simplification,we just consider the case of two classes.

Considering the size of sub-region may have an effect on classifier's performance, we divide the images ($256\times256$) into eight different scales: $2\times2$, $4\times4$, $8\times8$, $16\times16$, $32\times32$, $64\times64$, $128\times128$ and $256\times256$. And it can been seen from Fig. 6 that high order structure features have discriminative ability and when the sub-region size reaches $64\times64$ or $128\times128$, the classification accuracy  reach the best results. High performance can be is achieved by the proposed TOSF. However, a small sub-region size will also result in the decrease of deteriorate the classification accuracy. This is consistent with our intuition in that the proposed TOSS and FOSS are holistic scene representations for modeling the scene structure. In other words, high order structure features can be seen as novel holistic feature descriptor. \\
\\
\begin{figure}[bhpt]
 \centering
 \includegraphics[width=2.8in]{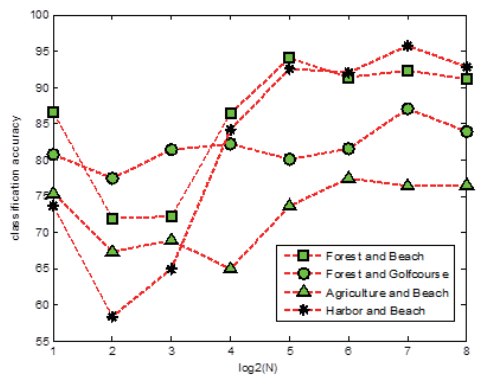}
 \vspace{-1em}
\caption {Classification accuracy for different sub-region size with TOSF.}
\vspace{-3em}
\end{figure}
\section{CONCLUSION}
In this paper, we have proposed higher order structure statistics (TOSS and FOSS) as novel structure descriptors and obtain a better result in image reconstruction. Compared with the phase spectrum, TOSS and FOSS possess fine property, that is, the surface of them can be conceptualized in a more unitary form than phase spectrum. Then, we apply higher order structure statistics to extract higher order structure features (TOSF and FOSF). It demonstrates the effectiveness of these features scene image categorization, and whether these features can maintain effectiveness in the case of multi-classes is worthy of further research.


\section{Acknowledgement}
\label{sec:typestyle}
This work was supported in part by the National Basic Research
Program of China(2012CB719903).
\section{REFERENCES}
\label{sec:refs}
\vspace{-2em}
\renewcommand\refname{}
\bibliographystyle{IEEEbib}
\bibliography{research}

\end{document}